\documentclass[letterpaper, 10 pt, conference]{ieeeconf}  %
\IEEEoverridecommandlockouts
\usepackage{cite}
\usepackage{amsmath,amssymb,amsfonts}
\usepackage{algorithmic}
\usepackage{graphicx}
\usepackage{textcomp}
\usepackage{xcolor}
\usepackage{graphicx} 
\usepackage{amsmath}
\usepackage[flushleft]{threeparttable}
\usepackage{afterpage}
\usepackage{amssymb}  
\usepackage{amsfonts}
\usepackage{algorithmic}
\usepackage{array}
\usepackage{float}
\usepackage{textcomp}
\usepackage{stfloats}
\usepackage{url}
\usepackage{verbatim}
\usepackage{graphicx}
\usepackage[ruled, vlined, linesnumbered]{algorithm2e}
\usepackage{cite}
\usepackage{amssymb,amsfonts,bm}
\usepackage{amsmath,tikz}
\usepackage{color}
\usepackage{mathtools}
\usepackage{hyperref}
\hypersetup{
    colorlinks=true,
    linkcolor=black,
    filecolor=magenta,    
    citecolor=black,
    urlcolor=blue,
    }
\usepackage{rotating}
\usepackage{blkarray}
\usepackage{physics}
\usepackage{bm}
\usepackage[normalem]{ulem} 
\usepackage{graphicx}
\usepackage{booktabs}
\usepackage[linesnumbered,ruled,vlined]{algorithm2e}
\usepackage{multirow}

\usepackage{geometry}
\newgeometry{top=0.8in, bottom=0.6in, left=0.667in, right=0.667in}

\def\BibTeX{{\rm B\kern-.05em{\sc i\kern-.025em b}\kern-.08em
    T\kern-.1667em\lower.7ex\hbox{E}\kern-.125emX}}

\def\tf{\bm{T}}         
\def\pos{\mathbf{p}}    
\def\te{\mathrm{text}}
\def\entity{\mathcal{E}}
\def\ltem{\mathcal{M}}

\title{\LARGE \bf
Robust Loop Closure by Textual Cues in Challenging Environments
}

\author{
    Tongxing Jin$^\dagger$,
    Thien-Minh Nguyen$^\dagger$, \IEEEmembership{Member,~IEEE},
    Xinhang Xu,
    Yizhuo Yang,\\
    Shenghai Yuan$^*$, \IEEEmembership{Member,~IEEE},
    Jianping Li,
    Lihua Xie, \IEEEmembership{Fellow,~IEEE}
\thanks{The authors are with the School of Electrical and Electronic Engineering, Nanyang Technological University, 50 Nanyang Avenue, Singapore 639798. Emails: \texttt{\{tongxing.jin, thienminh.nguyen, xinhang.xu, yang0670, shyuan, jianping.li, elhxie\}@ntu.edu.sg}}.
\thanks{This research is supported by the National Research Foundation, Singapore, under its Medium-Sized Center for Advanced Robotics Technology Innovation (CARTIN).}
\thanks{$^\dagger$ Joint first authors with equal contribution.}
}

\begin{document}

\maketitle

\begin{abstract}
Loop closure is an important task in robot navigation. However, existing methods mostly rely on some implicit or heuristic features of the environment, which can still fail to work in common environments such as corridors, tunnels, and warehouses. Indeed, navigating in such featureless, degenerative, and repetitive (FDR) environments would also pose a significant challenge even for humans, but explicit text cues in the surroundings often provide the best assistance.
This inspires us to propose a multi-modal loop closure method based on explicit human-readable textual cues in FDR environments. Specifically, our approach first extracts scene text entities based on Optical Character Recognition (OCR), then creates a \textit{local} map of text cues based on accurate LiDAR odometry and finally identifies loop closure events by a graph-theoretic scheme.
Experiment results demonstrate that this approach has superior performance over existing methods that rely solely on visual and LiDAR sensors.
To benefit the community, we release the source code and datasets at \url{https://github.com/TongxingJin/TXTLCD}.

\end{abstract}


\section{Introduction}

In recent years, LiDAR-inertial odometry (LIO) has become a pillarstone in the field of mobile robotics \cite{xu2022fast, lin2022r, nguyen2023slict}. 
Notably, the newer Livox Mid-360 3D LiDAR now costs similarly to the Intel Realsense D455 camera, offering a wider field of view, longer range, and improved accuracy.
In terms of continuous localization, LiDAR-based methods have decisively demonstrated better accuracy and robustness over conventional visual SLAM and eliminate the need for visual SLAM in most applications, as reflected in the Hilti SLAM challenge \cite{zhang2022hilti}.
However, LiDAR-based loop closure detection (LCD) methods, such as Stable Triangle Descriptor (STD)\cite{10160413}, Scan Context (SC)\cite{kim2018scan}, and Intensity Scan Context (ISC)\cite{wang2020intensity}, often struggle to find accurate matches in degenerative and repetitive environments. While visual-based LCD \cite{galvez2012bags, arandjelovic2016netvlad, izquierdo2024optimal} offers larger feature descriptor dimensions, most of them are sensitive to illumination and viewpoint variance.

Despite dense feature representation, visual-based LCD failures still occur. The LCD challenge arises when dealing with featureless, degenerative, and repetitive  (FDR) environments with computational constraints. 
There exists a gap for an efficient, simple, and intuitive LCD solution for LIO in FDR environments, mirroring human-like processes.

\begin{figure}
  \centering
   \includegraphics[width=0.98\linewidth]{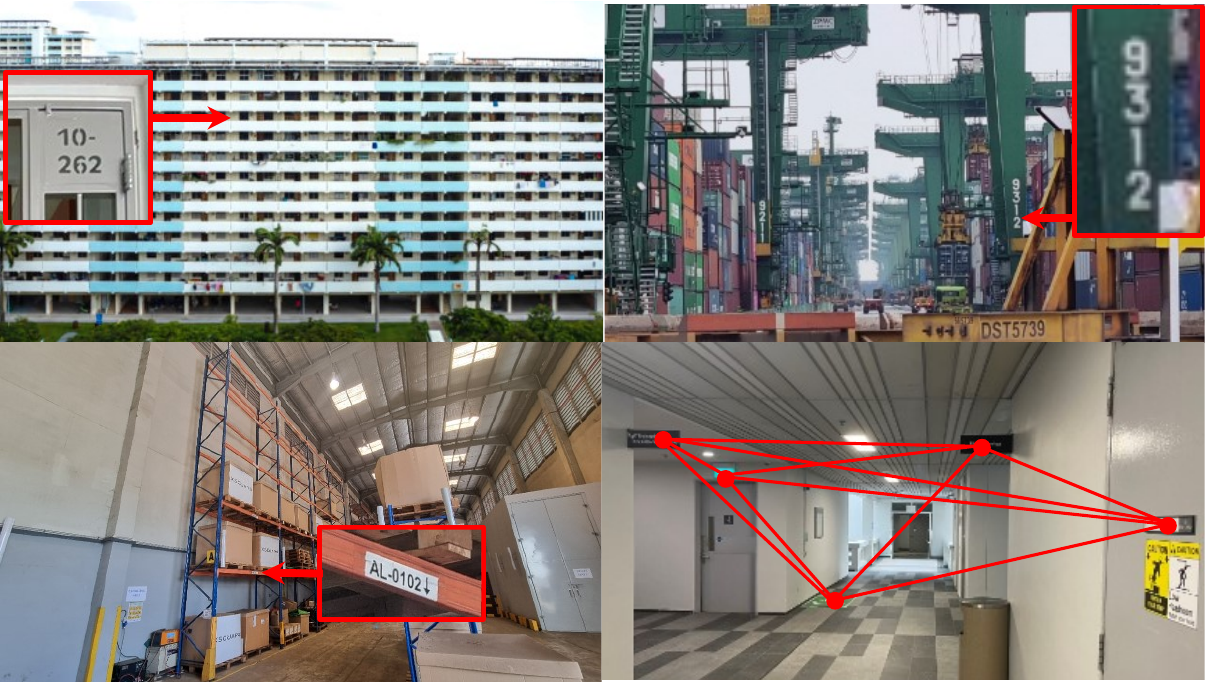}
   \caption{Examples of common FDR scenes, where humans naturally navigate using readable textual signs and their spatial arrangements. This inspires us to use text cues for global location understanding.}
   \label{front}
\end{figure}

The research presented in this paper draws inspiration from the observation that humans often rely on textual cues within their environments to determine their location. Indeed, these text cues are often designed with the clear intention of helping humans to navigate in FDR environments (Fig. \ref{front}) and can come in many forms, such as wayfinding signs, nameplates, and other forms of language-based signage. 
TextSLAM\cite{li2023textslam, li2020textslam} is the first that tightly integrates scene texts into a visual SLAM pipeline and demonstrates its effectiveness in a text-rich commercial plaza. In contrast to their approach, our method further takes advantage of the spatial structure of scene texts to verify the authenticity of loop closures, while effectively operating in environments with moderate text density, typical of most real-world scenarios.

Based on this inspiration, we propose a multi-modal (MM) loop closure solution that leverages scene text cues in FDR scenes.
Specifically, we employ the mature visual Optical Character Recognition (OCR) technology to detect scene text entities present in the vicinity of the current location, and then, based on low-drift LIO, create a \textit{local text entities map} (LTEM) to encode the special spatial arrangements of these texts, which can be used as a token to verify the authenticity of the candidate loop closures. Loop closure constraints will be introduced to the backend pose graph to enhance the robustness and accuracy of state estimation in typical FDR environments. The contributions of our work can be summarized as follows:

\begin{enumerate}

  \item We introduce a novel text entity representation, estimation, and management method by fusing LiDAR and visual data, which supports efficient loop closure retrieval and alignment.

  \item We propose an association scheme for observations of the same text entity, which are then used to create loop closures and improve state estimation accuracy. In particular, we employ a graph-theoretic method to identify the authenticity of candidate loop closures.
  
  \item We integrate our method with LiDAR odometry to form a SLAM framework and conduct extensive experiments to demonstrate its competitive performance compared with state-of-the-art (SOTA) methods.

  \item We release our source code and survey-graded high-precision dataset for the benefit of the community.
\end{enumerate}

\section{Related work}
Visual and LiDAR fusion for global localization is a common problem in perception tasks, which has been tackled in various prior research.

Traditionally, global localization can be achieved by visual odometry or SLAM approach. However, visual approaches often lack robustness when dealing with featureless areas, illumination changes and distant objects \cite{yuan2021survey}. Often, visual factors have to be complemented by other factors, such as IMU or UWB \cite{nguyen2023vr}, to increase efficiency and robustness. Recently, LiDAR approaches have become the mainstream for front-end odometry estimation as LIO \cite{xu2022fast, nguyen2023slict} consistently yields superior results when compared with most real-time method that fuses with vision. With the emergence of new low-cost LiDAR-based solutions, the visual-based approach has dropped in popularity for robotics localization \cite{zhang2022hilti}.

Another way to achieve global localization is by LCD. Traditionally, the visual-based method was mainstream with handcrafted features.
DBOW2 \cite{galvez2012bags} improved real-time LCD using binary visual word models based on BRIEF features\cite{calonder2010brief}. In recent decades, learning-based methods\cite{arandjelovic2016netvlad} have dominated LCD due to their better performance in handling viewpoint and appearance variations. As an extension of \cite{arandjelovic2016netvlad}, an optimal transport aggregation-based visual place recognition model was proposed in \cite{izquierdo2024optimal}, achieving SOTA results on many benchmarks.
However, due to limited geometric understanding, FDR environments, and changes in illumination, visual-based LCD is still far from perfect.

Recently, LiDAR-based LCD has been extensively explored in field robotics for precise geometric measurements and illumination invariance.
The SC series is currently considered the most popular method for LiDAR-based LCD \cite{kim2018scan, kim2021scan}, whose main idea is to employ projection and space partitioning to encode the entire point cloud, and later works have improved on this idea by integrating intensity \cite{wang2020intensity} and semantic information \cite{li2021ssc}. However, this series of methods cannot estimate the full SE3 relative pose between candidate frames and rely on odometry poses to reject false loops, making them vulnerable to significant odometry drift.

STD \cite{10160413} proposes creating a triangle-based descriptor by aggregating local point features, using the length of each side as a key in a hash table for finding loop closure candidates through a voting scheme. Recent work, known as Binary Triangle Combined (BTC) \cite{10388464}, combines STD with binary patterns to improve speed and viewpoint invariance. BTC is currently in early access and not yet available for open-source verification. However, these methods struggle with FDR scenes, where similar space shape, intensity, and semantics can cause ambiguity in loop closure. 


TextSLAM\cite{li2023textslam, li2020textslam} is the first to tightly integrate scene texts into a point-based visual SLAM framework.
It selects the top ten historical keyframes that observe the most co-visible text objects as candidates for loop closure. The co-visibility requirement, which necessitates multiple text objects being visible in the loop closure frames, limits its applicability in real-world environments. To overcome this limitation, we create a local text entities map with the help of low-drift LIO and check the authenticity of candidate loop closures using the spatial arrangements of scene texts, thus making it effective in more common scenarios with moderate text densities.

\section{Methodology}

\begin{figure*}
    \centering
   \includegraphics[width=0.85\linewidth]{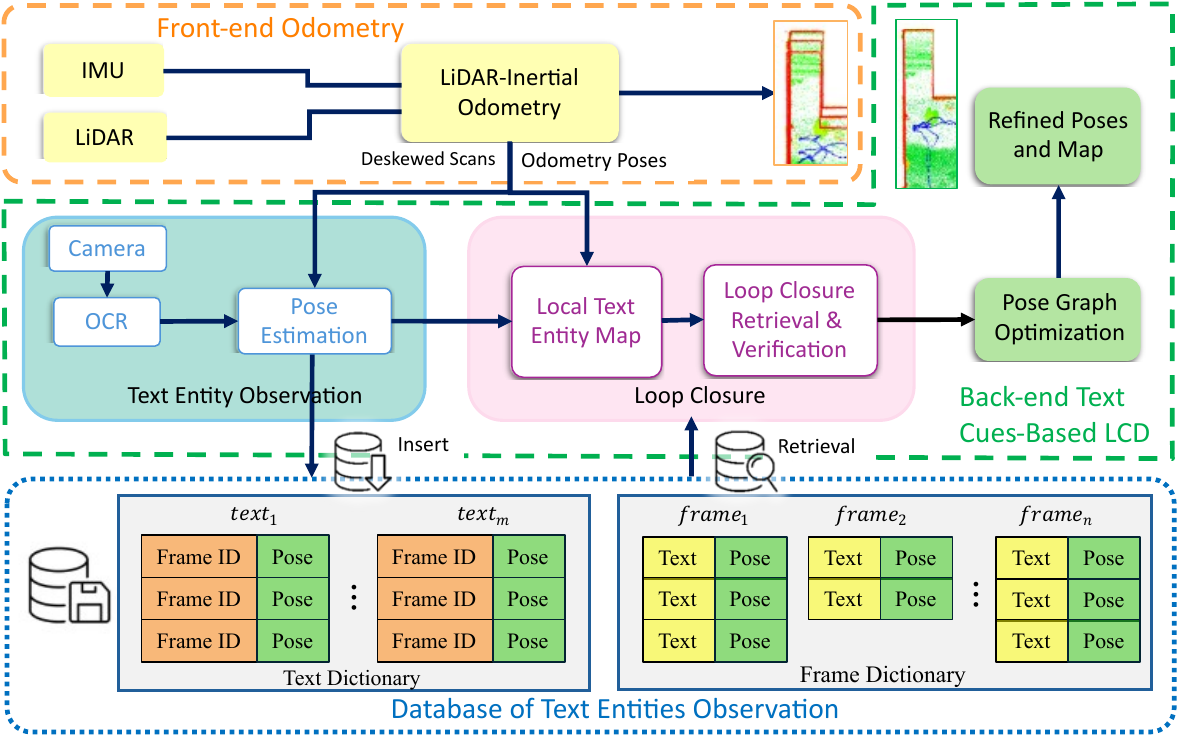}
   \caption{Pipeline of our text cue-based loop closure. Camera and LiDAR data are fused to estimate text entity poses and create local text entity maps that encode the specific arrangement of the scene texts. A novel graph-theoretic scheme is applied to verify the authenticity of candidate loop closures retrieved from the online database, and pose graph optimization is performed whenever a new loop is closed to mitigate cumulative odometry drift and ensure the consistency of global maps.}
   \label{workflow}
\end{figure*}

In this section, we describe the process of representing scene texts as text entities with content and pose attributes and observing them in a LiDAR frame. We then explain the principle of creating association relationships between text observations and identifying the authenticity of candidate loop closures using a graph-theoretic approach. The workflow of our method is depicted in Fig.~\ref{workflow}.

\textit{Notations}:
We define four main coordinate frames: the world frame $W$, LiDAR frame $L$, and camera frame $C$. We use $\tf_{L_t}^W$ to represent the SE3 pose of LiDAR in the world frame at timestamp $t$. For simplicity, we may omit the superscript $W$ for the world frame and rewrite them as $\tf_{L_t}$ and $\tf_L$. In addition, LiDAR poses $\{\tf_{L_t}\}_{t=t_0}^{t_n}$
are adopted as the nodes in back-end pose graph optimization. Similarly, $\tf_C^L$ will be used to express the extrinsic parameters between camera and LiDAR, $\tf_{\te}^C$ and $\tf_{\te}^L$ are text entity poses expressed in camera and LiDAR frames respectively.

\subsection{Text Entity Observation}
We abstract scene texts into text entities that contain two attributes: text contents and SE3 poses. Text contents refer to the text string that could be achieved by OCR, while pose observation is achieved by the fusion of camera and LiDAR measurements.
\subsubsection{Text Content Interpretation}
OCR is a mature technique that first locates the text regions in an image in the form of polygons and then converts the regions of interest into readable text content. In our implementation, we adopt AttentionOCR\cite{zhang2019feasible} to extract scene texts, which provides confidence scores to assist in filtering out unreliable recognition results.


\subsubsection{Text Entity Representation}
\begin{figure}
  \centering
   \includegraphics[width=0.9\linewidth]{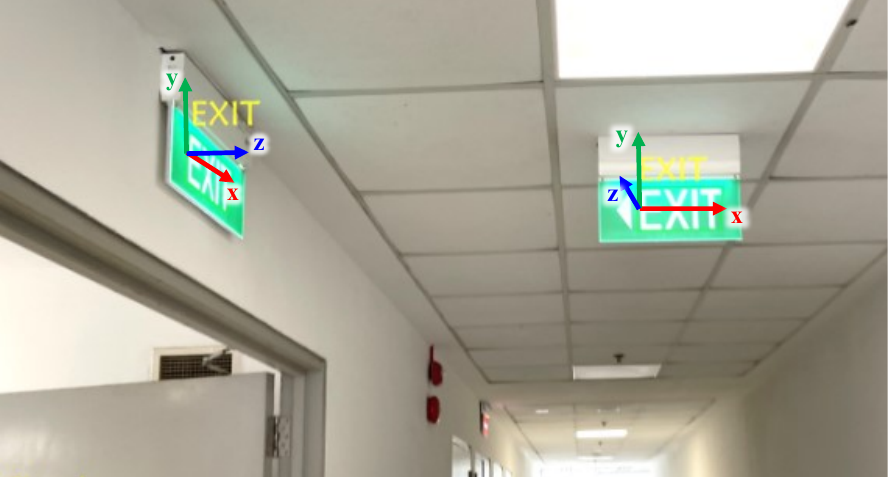}
   \caption{Illustrations of Text Entity Representation.}
   \label{fig: exit}
\end{figure}

Inspired from TextSLAM\cite{li2023textslam, li2020textslam}, it is reasonable to assume that scene text entities are typically situated on flat surfaces or local planes. Examples include notices on bulletin boards, room numbers, nameplates on fire-fighting facilities, and emergency exit signs. As shown in Fig.~\ref{fig: exit}, we define the midpoint of the left edge of the scene text region as the origin of the text entity. The x-axis points towards the midpoint of the right edge of the text, the z-axis aligns with the normal direction of the local plane and points towards the camera, and the y-axis is determined by the right-hand rule.

\subsubsection{Pose Estimation} To estimate the SE3 pose of the text entity in the camera frame, we first accumulate LiDAR scans in the past one second into a local point cloud map and cast it into the camera frame by extrinsic parameters between LiDAR and camera:
\begin{align}
\pos^C = \tf_L^C \circ \pos^L=[p^C_{x},\ p^C_{y},\ p^C_{z}]^\top,
\end{align}
where $\pos^L$ is a LiDAR point coordinate in the LiDAR frame, $\tf_L^C$ is the extrinsic parameter between LiDAR and the camera frame and $\pos^C$ is the point coordinate in the camera frame. Then, the LiDAR points will be further projected into image coordinates: 
\begin{equation}
    \begin{bmatrix}
        u&v&1
    \end{bmatrix}^\top
    =\mathbf{K}\left[\frac{p^C_x}{p^C_z},\ \frac{p^C_y}{p^C_z},\ 1\right]^\top,
\end{equation}
where K is the intrinsic matrix of the camera, and $[u, v]^\top$ is the pixel coordinate that the LiDAR point falls in.  

As the scene text is typically affixed to a local plane, the plane parameters in the camera frame can be estimated through RANSAC on the set of points detected within the region where the text is present. We represent the plane in the camera frame as:
\begin{equation}
    \mathbf{n}^\top\pos+d=0, \label{entity1}
\end{equation}
where $\mathbf{n}$ is the normal of the plane, $\pos$ is any point that lies on the plane, and $d$ is the distance from the camera's optical center to the plane.
Given plane parameters $(\bm{n}, d)$ and the projection coordinates $[u,v]$ of a point $\pos^C$, the depth of the point can be recovered as follows:
\begin{equation}
    p^C_z=-\frac{d}{\mathbf{n}^\top\mathbf{K}^{-1}[u,v,1]^\top}, \label{entity2}
\end{equation}
and its position $\mathbf{p}^C$ could be expressed by:
\begin{equation}
    \pos^C = p^C_z \mathbf{K}^{-1}[u,v,1]^\top, \label{entity3}
\end{equation}

Each text entity detected by OCR comes with a bounding box. 
We denote the midpoint of the left side and the right side of the bounding box as $\bm{p}_l^C$ and $\bm{p}_r^C$, respectively.
We chose $\bm{p}_l^C$ as the position of the text entity, and $\bm{n}_x \triangleq \frac{\bm{p}_r^C - \bm{p}_l^C}{\norm{\bm{p}_r^C - \bm{p}_l^C}}$ the unit vector of the $x$-axis. The pose matrix of the text entity is therefore defined as:
\begin{equation}
    \tf_{\te}^C=
    \begin{bmatrix}
        \mathbf{n}_x & \mathbf{n} \times \mathbf{n}_x & \mathbf{n} & \pos_{l}^C\\
        0&0&0&1
    \end{bmatrix}.
\end{equation}

Since the camera and LiDAR are different modal sensors and are triggered at different time points, the text entity will be further anchored into the latest LiDAR frame at timestamp $t_i$ just before image timestamp $t_j$, and its SE3 pose in LiDAR frame will be expressed as:
\begin{equation}
    \tf^{L_i}_\te=\tf_{L_j}^{L_i}\tf^L_C\tf_{\te}^{C_j},
\end{equation}
\begin{equation}
    \tf_{L_j}^{L_i}={\mathrm{interpolate}}\left({\tf_{L_i}^{-1}}{\tf_{L_k}}, \frac{t_j-t_i}{t_k-t_i}\right),
\end{equation}
where $t_i$ and $t_k$ are the two nearest LiDAR timestamps just before and after image timestamp $t_j$, respectively. $\mathrm{interpolate}(\tf, s)$ is the linear interpolation between the identity transform and $\tf$ by a factor $s \in (0, 1)$; $\tf_{\te}^{L_i}$ is the SE3 pose of text entity in its anchored LiDAR frame. For simplicity, henceforth, we will only deal with the text entity's pose relative to the LiDAR frame $\tf^L_\te$.


\subsection{Text Observations Management}
To support efficient loop closure storage, retrieval and alignment, we keep all of the historical text entity observations in a \textit{database of text entities observation}, which consists of a \textit{text dictionary} and a \textit{frame dictionary} implemented by a hash map (Fig.~\ref{workflow}). The \textit{text dictionary} uses text strings as keys, indices of all LiDAR frames observing the text content, as well as their estimated text entity poses, as value, and this enables fast retrieval of candidate frames that observe a certain text content. The \textit{frame dictionary} utilizes frame indices as keys, contents and estimated poses of all observed text entities in this frame as value, facilitating the creation of local text entities map near the candidate frame. 

\subsection{Text Entity Based Loop Closure Detection and Alignment}
There is a diverse array of scene texts found across various environments, offering insights into the function and location of the associated entity. Unlike QR codes or other landmarks, scene text holds the advantage of not requiring specialized deployment and can seamlessly integrate with human navigation. We classify scene texts in two categories: \textit{ID texts} and \textit{generic texts}, where ID texts are address-like texts that help us identify specific rooms or objects, and generic texts are everything else, for example, \textit{EXIT}, \textit{DANGER}, \textit{POWER}. Based on the text entities, we apply different loop closure detection strategies.

\subsubsection{ID Texts}
ID texts refer to texts that follow special conventions designed by humans to identify the specific objects within a building or map. For instance, \textit{S1-B4c-14} denotes building S1, the fourth basement level (a below-ground floor), section c, and room 14, while \textit{S2-B3c-AHU3} encodes building S2, the third basement level, section c, and \textit{Air Handling Unit 3}. Such texts can be picked out according to the predefined pattern of the application environment.

ID texts are usually designed to be exclusive, such as door numbers or equipment numbers, thus repeated detection of the same ID text content at different times indicates a high possibility of a loop closure. The relative pose prior $\bar{\tf}^{L_i}_{L_j} $ between the current and candidate loop closure poses $\tf_{L_i}$ and $\tf_{L_j}$ is calculated as following,

\begin{equation}
    \bar{\tf}^{L_i}_{L_j} = {\tf_{\te}^{L_i}}({\tf_{\te}^{L_j}})^{-1}.\label{eq: relative_pose}
\end{equation}

However, the ID texts can also have multiple instances at different locations, for example, a room may have multiple doors with the same number. Thus, we use an ICP check to rule out the wrong loop candidates from such cases. Moreover, ICP can provide a more precise relative pose prior \eqref{eq: relative_pose}, which is beneficial to the global pose-graph optimization task.

\subsubsection{Generic Texts}
In general, a substantial portion of scene texts do not indicate exclusive location information and can appear multiple times within a scene, for example, \textit{EXIT}, \textit{No Parking}, and \textit{STOP}. The association of such text entities can be ambiguous. 
To address this issue, we create a \textit{local text entity map} (LTEM) by aggregating text entities in the vicinity of the current location. Such an LTEM encodes the spatial arrangements and can be used as a token of the current pose to verify the authenticity of candidate loops with other poses, which we shall explain below.

Specifically, an LTEM is a set of all text entities, including both ID texts and generic texts, observed by a set of LiDAR odometry poses. Assuming that at the \textit{current pose} $\tf_c$ (the subscript $c$ is for \textit{current}), we observe a text entity $\entity_c$. We define $\ltem_c$ as the LTEM that contains all of the text entities observed by the continuous poses $\mathcal{T}_c=\{\tf_{c-w} \dots \tf_{c}\}$, where $\tf_{c-w}$ is the earliest pose within a certain distance $d$ from $\tf_c$. Note that $\ltem_c$ may contain other text entities whose \textit{content}, i.e., the text string, is different from $\entity_c$.

\begin{figure}
  \centering
   \includegraphics[width=1\linewidth]{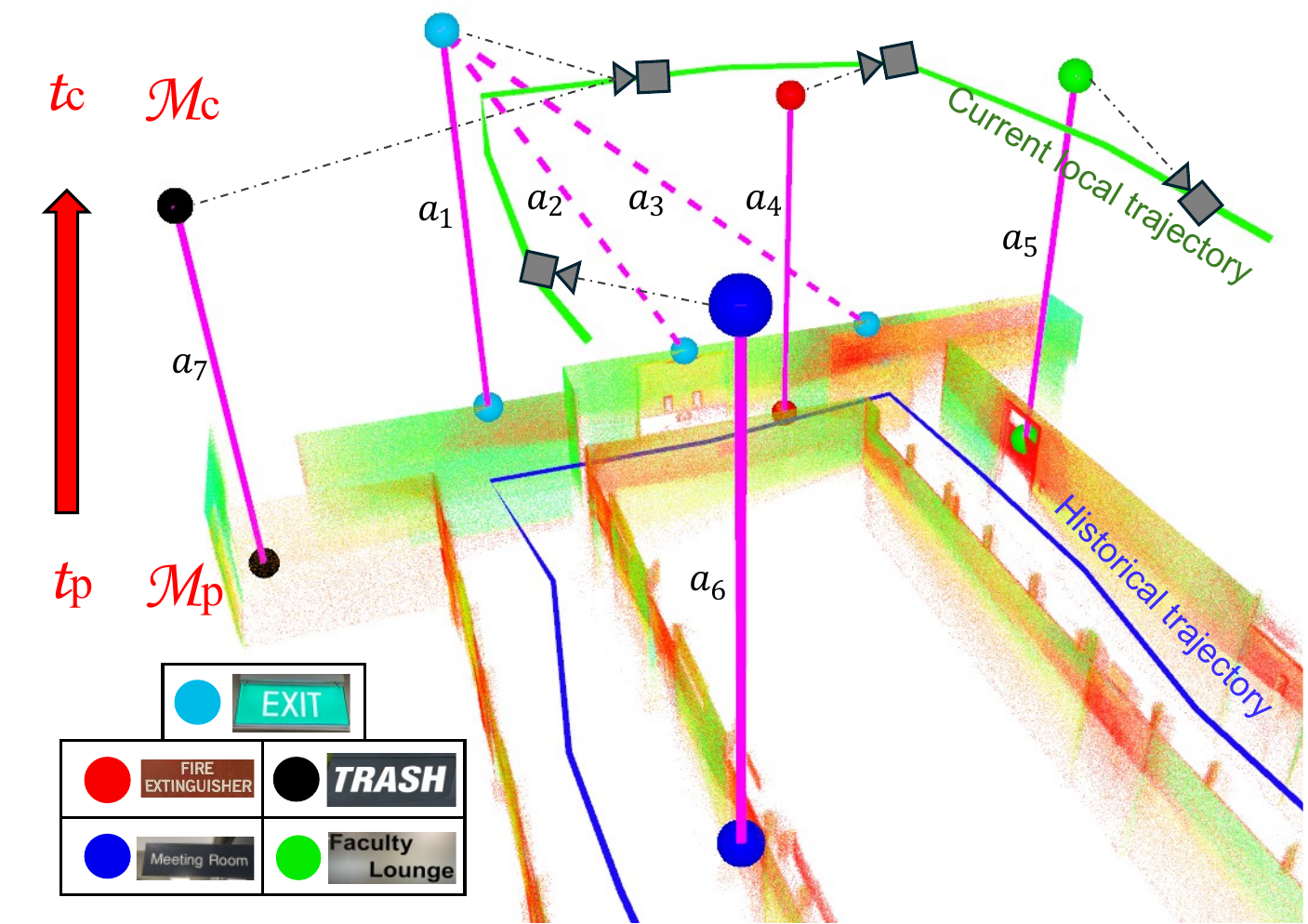}
   \caption{Putative associations between two LTEMs. LTEM $\ltem_c$ and $\ltem_p$ contain a set of text entities observed by the continuous LiDAR poses $\mathcal{T}_c$ (green trajectory) and $\mathcal{T}_p$ (blue trajectory), respectively. Text entities with identical text contents are represented by balls of the same color, connected by purple lines to indicate putative associations between the two LTEMs. The only \textit{EXIT} in $\ltem_c$ is associated with three different entities in $\ltem_p$. While association $a_1$ is the exclusively correct one, $a_2$ and $a_3$ (denoted by dashed purple line) should be rejected by our graph-theoretic loop closure verification method.}
   \label{fig: association}
\end{figure}

Then, we use the content of $\entity_c$ to search for all past poses that see a text entity with the same content from \textit{text dictionary}. Let us denote the set of these poses as $\mathcal{T}$.
For each \textit{candidate previous pose} $\tf_p \in \mathcal{T}$, we denote $\entity_p$ as the text entity with the same content with $\entity_c$ observed by $\tf_p$. We then construct an LTEM of all text entities observed by the continuous poses $\mathcal{T}_p=\{\tf_{p-w} \dots \tf_{p+v}\}$, where $\tf_{p-w}$ and $\tf_{p+v}$ are the earliest and latest pose within the same distance $d$ from $\tf_{p}$ respectively. We denote this LTEM as $\ltem_p$ (Fig. \ref{fig: association}).

Given $\ltem_c$ and $\ltem_p$, we will construct an association relationships set $\mathcal{A} \triangleq \{a_i,\dots\}=\{(\entity_i^c,\entity_i^p), \dots\}$, where $\entity_i^c \in \ltem_c$, $\entity_i^p \in \ltem_p$, and $\entity_i^c$, $\entity_i^p$ have the same text content. The set $\mathcal{A}$ is called the set of \textit{putative associations}. Apparently, $\mathcal{A}$ may contain inappropriate associations due to some repeated text contents. As shown in Fig.~\ref{fig: association}, associations $a_1$, $a_2$ and $a_3$ are mutually exclusive to each other, as they try to associate the same text entity from $\mathcal{M}_c$ to three different entities from $\mathcal{M}_p$.

\begin{figure}[t]
  \centering
   \includegraphics[width=0.85\linewidth]{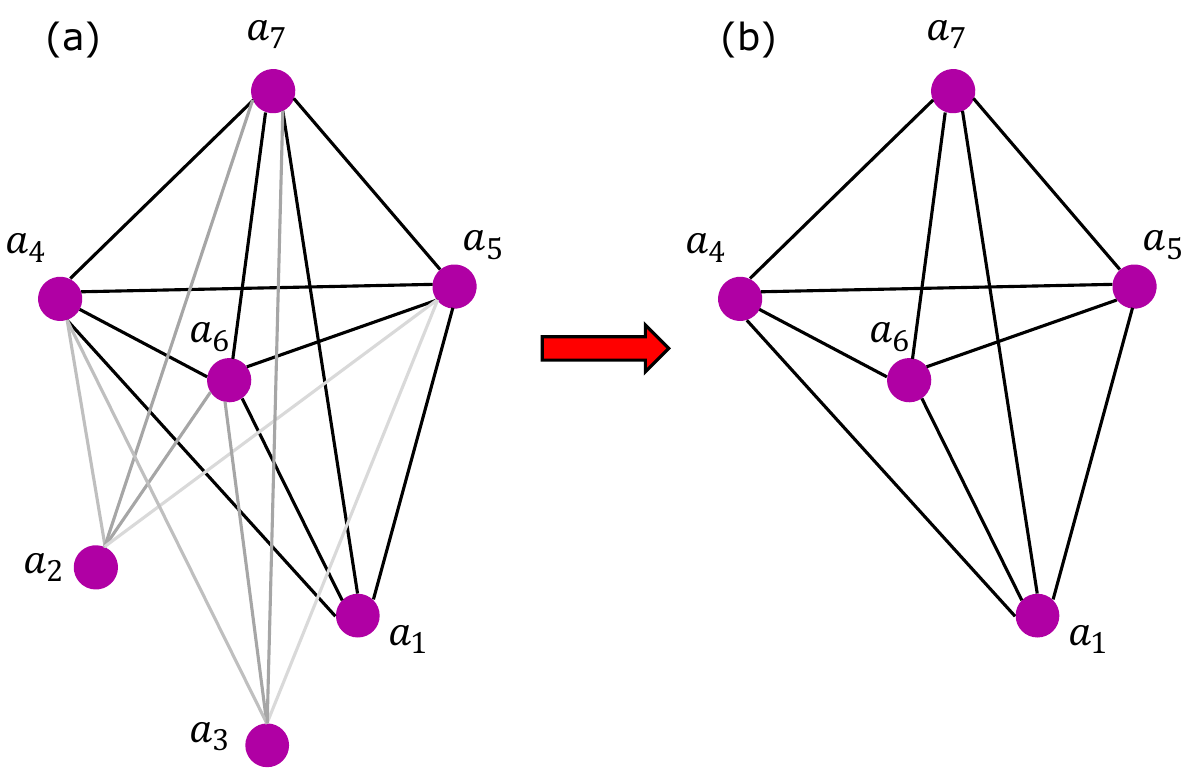}
   \caption{Consistency graph. The darkness of the lines signifies the geometrical consistency between the connected two nodes (putative associations).}
   \label{fig: consistency graph}
\end{figure}

The affinity relationships between these putative associations in $\mathcal{A}$ can be represented by a \textit{consistency graph} $\mathcal{G}$ as shown in Fig. \ref{fig: consistency graph} (a). Nodes in the consistency graph correspond to the putative associations in Fig.~\ref{fig: association}, and the connection between arbitrary two nodes $a_i$ and $a_j$ represents their compatibility, with the darkness of the line further signifying the \textit{geometrical consistency score} evaluated by:
\begin{equation}
    s_{i,j}(\abs{\norm{\pos_i - \pos_j} - \norm{\mathbf{q}_i - \mathbf{q}_j}}),
\end{equation}
where $\bm{p_i}$ and $\bm{q_i}$ are positions of the two text entities associated by $a_{i}$, $\bm{p_j}$ and $\bm{q_j}$ are positions of text entities associated by $a_{j}$, $\|\cdot\|$ means Euclidean norm of a vector, $s: \mathbb{R}\rightarrow[0,1]$ is a loss function subject to $s(0)=1$ and $s(x) = 0$ if $x>\varepsilon$ where $\varepsilon$ is the threshold value. This score indicates that the distance between two entities in one LTEM should match the distance between their counterparts in another LTEM, because the LiDAR odometry drift within an LTEM is negligible.

Next, we want to find a fully connected subgraph $\mathcal{G}^* \subset \mathcal{G}$ (Fig.~\ref{fig: consistency graph} (b)), as well as their nodes subset $\mathcal{A}^* \subset \mathcal{A}$, such that any pair of associations $a_i$ and $a_j$ in $\mathcal{A}^*$ are mutually consistent.
This problem is a variant of the maximum clique problem, and CLIPPER \cite{lusk2021clipper} formulates the problem as finding the densest subgraph $\mathcal{G}^*$. In this work, we employ CLIPPER to solve this problem.

Once the set $\mathcal{A}^*$ has been identified to have at least three elements and $(\entity_c, \entity_p) \in \mathcal{A}^*$, the two entities $\entity_c$ and $\entity_p$ can be used for constructing a relative pose constraint $\bar{\tf}^p_c$ for loop closure, similar to \eqref{eq: relative_pose}. We then iterate over all other poses in $\mathcal{T}$ to find all the possible loop closure constraints $\bar{\tf}^p_c$. The general process is summarized in Algorithm 1.

We note that in our multi-modal LCD and alignment scheme above, LIO output is tightly integrated with the visual detection information, first in the estimation of the text entity's pose and second in the construction of the LTEM. The high accuracy for short-term navigation of LIO is crucial for the performance of our method, and cannot be achieved with VIO, due to its poorer depth perception, and larger localization drift.

\begin{algorithm}
\caption{Association for repeatable text entities}
\KwData{Observed text entity $\entity_c$ from current pose $\tf_c$, database of text entities observation}
\KwResult{Loop closure text entities set $\{\entity_p,\dots\}$ with $\entity_c$}
Construct LTEM $\ltem_c$ by poses $\mathcal{T}_c=\{\tf_{c-w} \dots \tf_{c}\}$ using \textit{frame dictionary}\;
Retrieve the set of history poses $\mathcal{T}=\{\tf_p^1,\dots,\tf_p^n\}$ observing the same content as $\entity_c$ from \textit{text dictionary}\;
\For{$\tf_p \leftarrow \tf_p^1$ \KwTo $\tf_p^n$}{
    $\tf_p$ observed text entity $\entity_p$ which has the same content with $\entity_c$\;
    Construct LTEM $\mathcal{M}_p$ by poses $\mathcal{T}_p=\{\tf_{p-w} \dots \tf_{p+v}\}$ using \textit{frame dictionary}\;
    Create putative associations $\mathcal{A}$ between $\mathcal{M}_c$ and $\mathcal{M}_p$\;
    Find the geometrical consistent subset $\mathcal{A}^* \subset \mathcal{A}$\;
    \If{size of $\mathcal{A}^*$ $\geq 3$ and $(\entity_c, \entity_p) \in \mathcal{A}^*$}{Construct loop closure by $(\entity_c, \entity_p)$\;}
    }
\label{algorithm: 1}
\end{algorithm}

\section{Experiments}\label{Limitations}

In this section, we discuss the development of our datasets and the comparison with existing SOTA methods.
All of our experiments are conducted on a laptop with an Intel i7-10875H CPU @ 2.30GHz, and an NVIDIA GeForce RTX 2060 GPU.
A video summary of our experiments can be viewed on the project page listed in the abstract.

\subsection{Dataset and Experiment Setup}

To the best of our knowledge, few datasets are collected with attention to text cues. We note that in \cite{li2023textslam}, the text cues are sufficiently present, however since no LiDAR data is available, we cannot make a precise local text entity map. Another crucial requirement is that the camera and LiDAR FOVs need to have sufficient overlaps.
Due to these requirements, we find that no public dataset is available for text-based visual-LiDAR loop closure research.

\begin{figure*}
  \centering
   \includegraphics[width=0.90\linewidth]{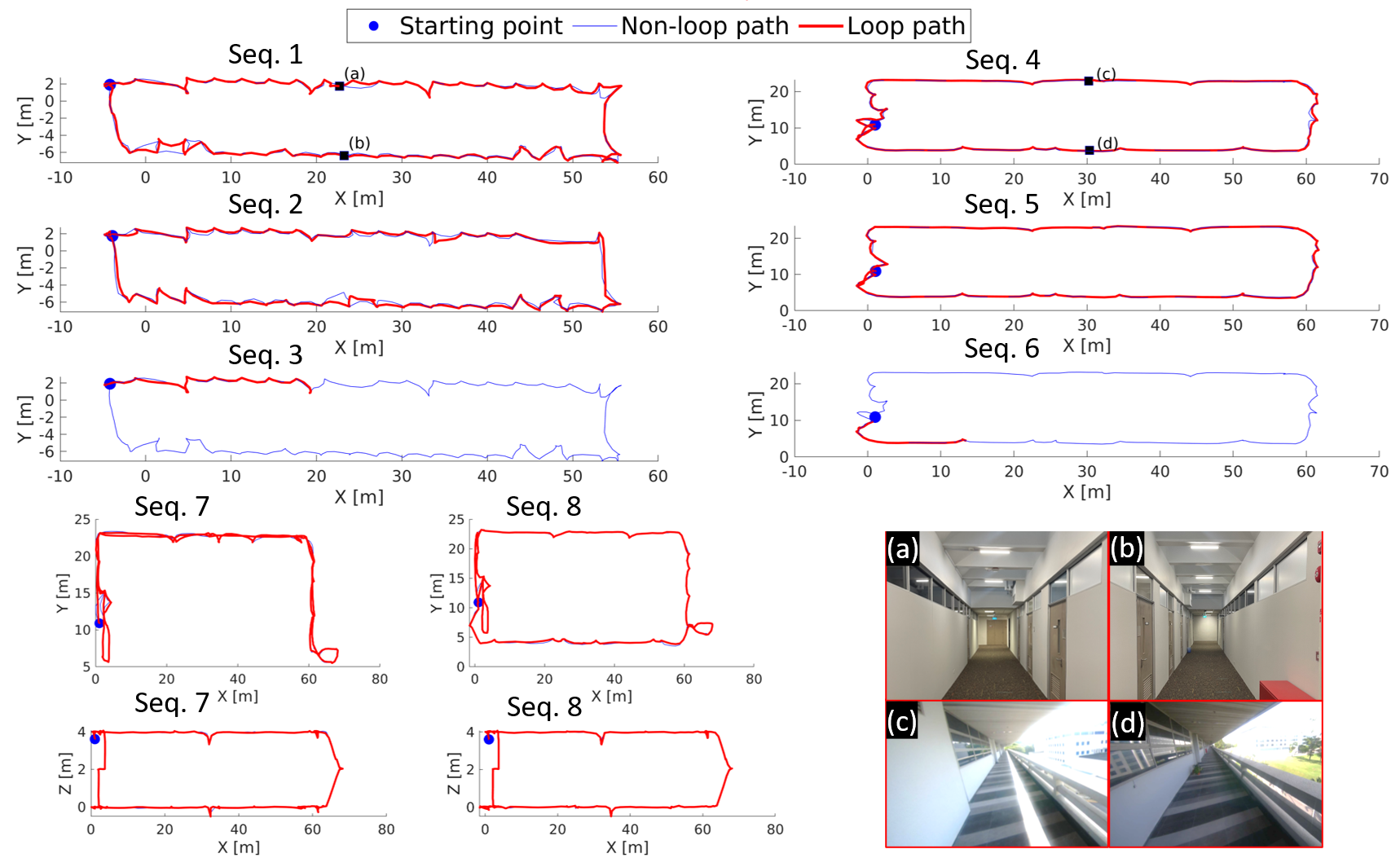}
   \caption{Illustrations of the trajectories in our dataset. The blue lines depict normal paths, while the red lines indicate locations of loop closure incidents (Sec. \ref{sec: loop closure incident}). Sequences 1, 2, 3 and 4, 5, 6 are captured on the same floor, whereas Sequences 7 and 8 traverse through different floors and vertical staircases. In (a), (b), (c), (d), we show very similar scenes at different corridors in Sequence 1 and 4.}
   \label{dataset}
\end{figure*}

To fill in this gap, we develop a high-quality dataset for multi-modal LCD in repetitive and degenerative scenes.
Our setup consists of a camera with $1920\times1080$ resolution, a Livox Mid360 LiDAR and its embedded IMU. For ground truth, we create high-precision prior point cloud maps of the environments with a Leica MS60 scanner, then register the LiDAR point clouds with these prior maps to obtain the ground truth trajectories, similar to \cite{ramezani2020newer, zhang2022hilti, mcd2024}. A total of 8 data sequences are collected from 3 different FDR scenes: an indoor corridor, a semi-outdoor corridor, and a cross-floor building with distances ranging from $200$ to $500$ meters. Their trajectories are illustrated in Fig.\ref{dataset}.

We compare our method with other popular open-source SOTA works, including SC \cite{kim2018scan}, ISC\cite{wang2020intensity}, and STD\cite{10160413}. To ensure fairness in the experiment, we integrate FAST-LIO2\cite{xu2022fast} with different loop closure methods to form complete SLAM systems for evaluation. We try to keep all of the parameters unchanged, except that ikdtree map size is set to be $100m\times100m$ with resolution $0.2m$, scans are downsampled at voxel resolution $0.1m$.

While our method is designed to address the LiDAR loop closure problem in FDR scenarios, we also input image sequences of our dataset into DBoW2\cite{galvez2012bags} and SALAD\cite{izquierdo2024optimal} to evaluate its recall and precision performance, as our method uses cameras to detect text.

\subsection{LCD Recall and Precision Analysis}

\subsubsection{True loop closure incidents} \label{sec: loop closure incident}

From the ground truth poses, we will evaluate each pose to determine if a loop closure detection should happen. Specifically, considering a pose $\tf_k$, we find the set $\mathcal{N}_k \triangleq \{\tf_p: \norm{\tf_k \boxminus \tf_p} < \tau \wedge S(\tf_k, \tf_p) > 10m,\ \forall p < k\}$, where $\tau$ is the Euclidean distance threshold which we set as 1.0m and 1.7m in our experiments, and travelled distance $S(\tf_k, \tf_p) \triangleq \sum_{i=p}^{k-1} \norm{\tf_{i+1} \boxminus \tf_i}$. If $\mathcal{N}_k \neq \varnothing$, then $\tf_k$ is marked as a \textit{loop closure pose}. 

For each loop closure method, we evaluate its recall rate and precision rate. The prediction of the method at pose $\tf_k$ can either be a TP, FP, TN, or FN based on the check of above $\mathcal{N}_k$. Hence, the recall rate is the ratio $\sum TP/(\sum TP + \sum FN)$ and the precision rate is the ratio $\sum TP/(\sum TP+\sum FP)$.

{
\begin{table*}

\newcommand{\tb}[1]{\textbf{#1}}
\newcommand{\ul}[1]{\underline{#1}}
\newcommand{\mc}[3]{\multicolumn{#1}{#2}{#3}}
\newcommand{\mr}[3]{\multirow{#1}{#2}{#3}}
\begin{threeparttable}
\caption{Recall and Precision Rates ([\%]), with different thresholds (1.0m / 1.7m)} \label{tab: precision and recall}

\begin{center}
\begin{tabular}{cccccccc|cccccc}
\hline\hline
\mr{3}{*}{\textbf{\#}}
& \mr{3}{*}{\tb{Scenario}}
& \mc{6}{c|}{\tb{Recall {[}\%{]}}}
& \mc{6}{c}{\tb{Precision {[}\%{]}}}
\\ \cline{3-14} 
&
& \mc{3}{c|}{\tb{LiDAR}}
& \mc{2}{c|}{\tb{Vision}}
& \tb{MM}
& \mc{3}{c|}{\tb{LiDAR}}
& \mc{2}{c|}{\tb{Vision}}
& \tb{MM}
\\ \cline{3-14}
&
& \mc{1}{c|}{\tb{SC}}
& \mc{1}{c|}{\tb{ISC}}
& \mc{1}{c|}{\tb{STD}}
& \mc{1}{c|}{\tb{DBoW2}}
& \mc{1}{c|}{\tb{SALAD}}
& \tb{Ours}
& \mc{1}{c|}{\tb{SC}}
& \mc{1}{c|}{\tb{ISC}}
& \mc{1}{c|}{\tb{STD}}
& \mc{1}{c|}{\tb{DBoW2}}
& \mc{1}{c|}{\tb{SALAD}}
& \tb{Ours} \\ \hline

1 & Indoor\_corridor & \tb{81}/\tb{82} & 2/2 & 7/8 & 18/18 & \ul{74}/\ul{73} & 60/59 & 48/51 & 86/86 & 57/66 & \ul{88}/\ul{88} & 67/67 & \tb{99}/\tb{100} \\
2 & Indoor\_corridor & \tb{53}/\tb{55} & 1/1 & 34/34 & 19/19 & 36/36 & \ul{51}/\ul{51} & 51/53 & 67/67 & 56/58 & \ul{81}/\ul{81} & 62/62 & \tb{97}/\tb{100} \\
3 & Indoor\_corridor\_part & 49/48 &  - / -  & 10/12 & 21/21 & \tb{76}/\tb{75} & \ul{58}/\ul{53} & 47/47 &  - / -  & 12/16 & \ul{69}/\ul{70} & 42/42 & \tb{100}/\tb{100} \\
4 & Semi-outdoor\_corridor & \ul{65}/\ul{65} & 2/2 & 5/5 & 43/43 & \tb{85}/\tb{84} & 49/49 & 67/71 & 80/80 & 80/85 & 37/39 & \ul{87}/\ul{92} & \tb{95}/\tb{100} \\
5 & Semi-outdoor\_corridor & \ul{45}/\ul{46} & 2/2 & 10/10 & 43/43 & \tb{74}/\tb{74} & 36/37 & 79/83 & 80/80 & 78/81 & 38/40 & \ul{84}/\ul{89} & \tb{97}/\tb{100} \\
6 & Semi-outdoor\_part & 21/21 &  - / -  & 23/20 & 32/28 & \tb{72}/\tb{62} & \ul{50}/\ul{45} & 23/26 &  - / -  & 32/32 & 10/10 & \ul{72}/\ul{76} & \tb{100}/\tb{100} \\
7 & Cross-floor\_building & \ul{45}/\ul{45} & 2/2 & 1/1 & 34/35 & \tb{80}/\tb{80} & 38/38 & 57/59 & 71/\ul{86} & 67/67 & 49/50 & \ul{82}/85 & \tb{100}/\tb{100} \\
8 & Cross-floor\_building & \ul{65}/\ul{65} & 2/2 & 6/7 & 41/42 & \tb{80}/\tb{79} & 39/37 & 52/54 & 83/83 & 76/82 & 75/80 & \ul{90}/\ul{92} & \tb{100}/\tb{100}\\

\hline
\end{tabular}
\end{center}
\begin{tablenotes}
      \item  ``MM" denotes Multi-Modality,   ``-" denotes no loop detected.
        Best results are \textbf{boldened}, and second-best results are \underline{underlined}.
    \end{tablenotes}
 \end{threeparttable}
\end{table*}
}

\subsubsection{Recall}

As shown in Tab. \ref{tab: precision and recall}, when $\tau = 1.0m$, SALAD achieves the best recall performance, exceeding $70$\% in most cases. Both our method and SC show competitive results, recalling more than $50$\% of loops in 4 sequences. The primary factor limiting our recall performance is the repeatability of the OCR module, which refers to its ability to consistently detect the same text string result across multiple observations of the same text entities.

The recall rates of ISC and STD are the lowest, usually lower than $10$\%, because they set more strict thresholds to confirm a true loop closure, leading to a relatively higher precision rate compared with SC.

In the context of loop closure, a higher recall rate may not indicate decisive superiority between two methods, since detecting at least one accurate loop for each repeated corridor is sufficient to significantly reduce odometry drift. Precision rate, however, is more critical because false loop closures can corrupt global pose estimation and map building.
In Sec. \ref{sec: pgo}, the adequacy of recalls can be verified by the effective reduction of absolute translation error when incorporating the detected loops into the pose-graph optimization stage.

\subsubsection{Precision}

As shown in Table~\ref{tab: precision and recall}, when $\tau = 1.0m$, SALAD demonstrates competitive performance with 5 second-best scores, while ISC stands out as the most effective pure LiDAR-based loop closure methods, achieving precision rates above $80$\% in 4 sequences. DBoW performs effectively in Sequences 1 and 2 (indoor corridors), with a precision rate above 80\%. However, the performance significantly drops to around 40\% in Sequence 4 and 5. Our FDR dataset is indeed challenging for DBoW as it erroneously generates many false loop closures between similar locations such as (a) and (b), (c) and (d) shown in Fig. \ref{dataset}, which could be difficult to distinguish even for humans.

\begin{figure}
  \centering
   \includegraphics[width=1\linewidth]{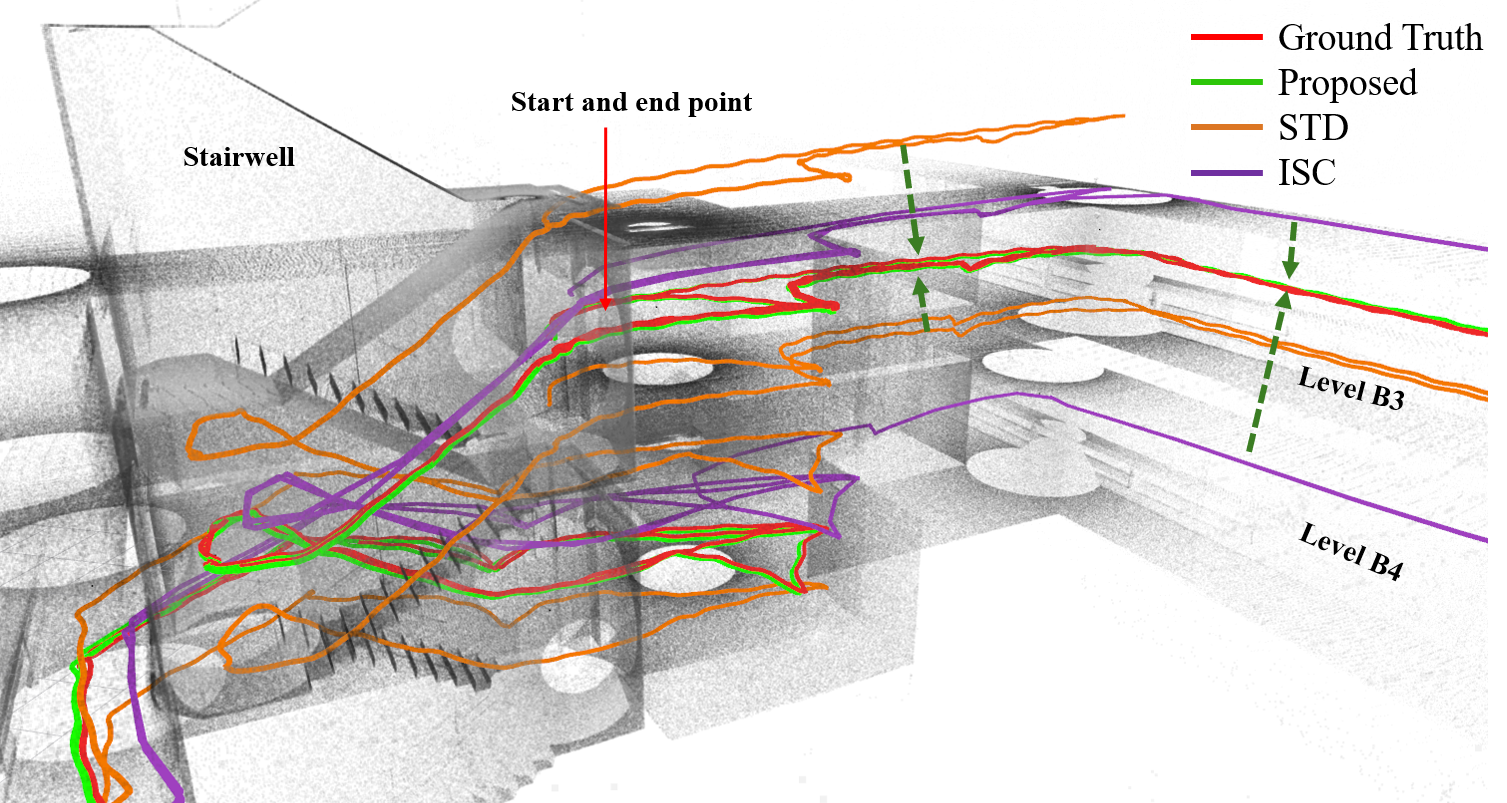}
   \caption{STD and ISC trajectories diverged significantly from the ground truth, with green dashed arrows indicating the convergence directions. In contrast, our trajectory remains consistently close to the ground truth.}
   \label{trajectory}
\end{figure}

The multi-floor building in Sequence 7 and 8 is a typical repetitive scene, as the layout of different floors is very similar, as shown in Fig. \ref{dataset} (c) and (d). Indeed, some humans may also find it hard to differentiate the corridors without the help of the text indicators.
One notable drawback of all comparative methods is their susceptibility to generating catastrophic false loop closures. Fig.~\ref{trajectory} shows part of the trajectories generated by several methods in Sequence 8. Both ISC and STD falsely associate frames from different floors as loop closures, causing their trajectories to deviate from the ground truth. However, our method still can use the room or equipment numbers as textual cues to differentiate different floors, avoiding the risk of forming erroneous loops.

Moreover, the precision of all SOTA methods deteriorates significantly in partially overlapped sequences compared to the fully overlapped ones, even though they are collected from the same environment, i.e., Sequence 3 vs 1--2 and Sequence 6 vs 4--5. This reveals their propensity to predict false loop closures, which becomes more pronounced when the trajectory overlap is relatively low.  

Comparatively, our method consistently achieves the best performance and obtains  
high precision rates of more than 95\% on all sequences
benefiting from our graph-theoretic loop closure identification scheme that effectively leverages the spatial arrangements of the text entities. The reason for the imperfect performance is that we set a tight threshold $\tau=1.0m$ to determine the incidence of loop closure as introduced in \ref{sec: loop closure incident}. If the threshold is relaxed slightly to $1.7m$, our method can achieve 100\% accuracy while maintaining the same recall rate as shown in Table \ref{tab: precision and recall}.

\subsection{Pose-Graph Optimization Error Evaluation} \label{sec: pgo}

We take FAST-LIO2 as the front-end odometry, and global pose graph optimization will be conducted while loop closure is detected. 
FDR environments pose significant challenges for visual odometry or SLAM methods like ORB-SLAM\cite{campos2021orb}, as many images are captured facing walls with nearly no features to extract or track continuously. Meanwhile, the visual method SALAD is not designed for loop closure and can not output relative pose estimation directly for the following global pose optimization. Therefore, we only analyze the pose error of different LiDAR LCD methods when integrated with FAST-LIO2 evaluated by EVO\cite{grupp2017evo}.

\begin{table}

\caption{Mean Absolute Translation Error [m]}
\begin{center}
    \begin{tabular}{cccccc}
    \hline
    \hline
       \#  & Pure Odometry & SC & ISC & STD & Ours\\ 
        \hline
       1  & \uline{0.118} & 0.179 & 0.526 & 1.725 & \textbf{0.07}\\
       2  & \uline{0.134} & 0.34 & 0.798 & 13.738 & \textbf{0.066}\\
       3  & 0.116 & \uline{0.08} & 0.666 & 2.48 & \textbf{0.068}\\
       4  & 0.401 & \uline{0.176} & 0.675 & 1.718 & \textbf{0.101}\\
       5  & 0.231 & \uline{0.111} & 0.479 & 0.359 & \textbf{0.043}\\
       6  & 0.365 & \uline{0.085} & 0.566 & 0.191 & \textbf{0.075}\\
       7  & 0.552 & \uline{0.193} & 0.774 & 1.967 & \textbf{0.164}\\
       8  & 0.167 & \uline{0.138} & 0.892 & 0.774 & \textbf{0.136}\\
    \hline
    \end{tabular}
    \label{APE}
\end{center}    
\end{table}

As shown in Table~\ref{APE}, by leveraging textual cues for loop closure, our method effectively minimizes the odometry drift across all datasets, consistently achieving the lowest mean translation error. In contrast, ISC and STD frequently report false loop closures, resulting in higher mean errors compared with odometry poses. The primary challenge in our dataset is its symmetric and repetitive layout, as shown in Fig.\ref{dataset} (a)-(d).

Although we avoid forming loop closures between neighboring poses within a traveled distance of $10m$ as explained in \ref{sec: loop closure incident}, SC can retrieve loops between poses whose travel distance is slightly larger than this threshold and introduce relative pose constraints between non-adjacent poses before a loop is closed, leading to a smaller mean translation error for SC compared with other methods.



\begin{figure}[t]
  \centering
   \includegraphics[width=1.0\linewidth]{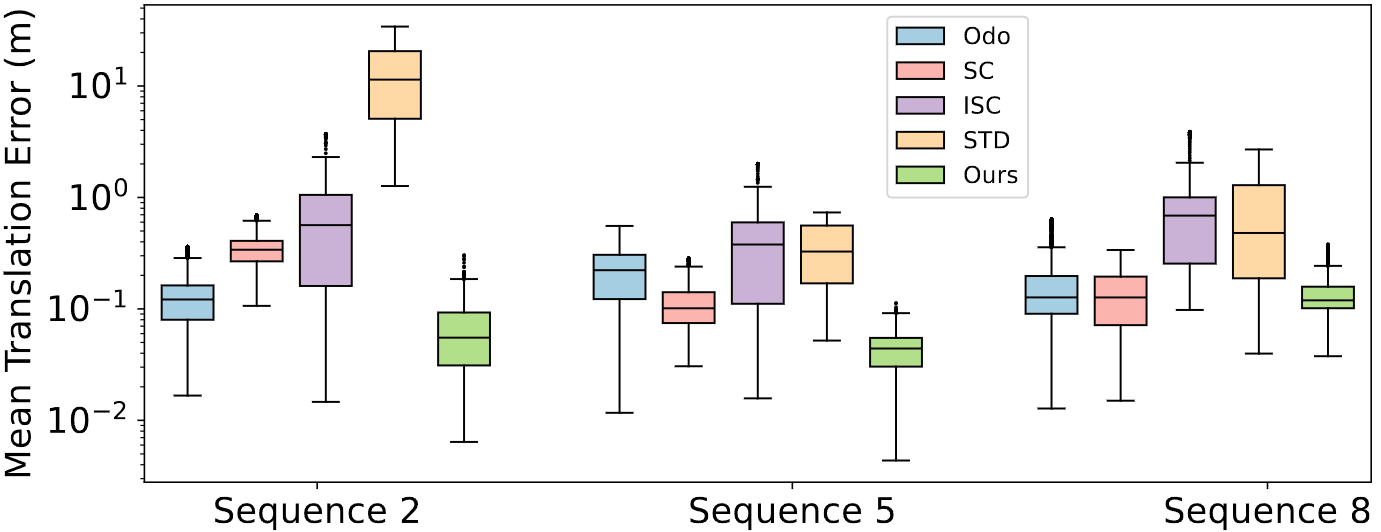}
   \caption{Translation error distribution.}
   \label{box}
\end{figure}

Besides the mean translation error, Fig.~\ref{box} further shows the error distribution of different methods in 3 sequences. It's evident that our method consistently achieves the lowest error upper bound compared with other methods, as all loops created by our method are authentic, and no erroneous constraints are introduced into the pose graph. Moreover, the spread of our localization error remains consistent across different sequences.

\subsection{Runtime Analysis}
We evaluate the time cost of different stages of our methods in Sequence 1, 4, and 7, respectively. The results, detailed in Table~\ref{runtime}, show that OCR is the most time-consuming part. However, it can be replaced with other OCR methods in the future.

\begin{table}
\caption{Mean Runtime [s]}
\begin{center}
    \begin{tabular}{ccccc}
    \hline
    \hline
        \# & OCR & Text Entities Extraction & Loop Closure & Total\\ 
         
        \hline
       1 & 0.603	& 0.034	& 0.087 & 0.724\\
       4 & 0.641	& 0.033	& 0.081 & 0.755\\
       7 & 0.597	& 0.036	& 0.070 & 0.703\\
    \hline
    \end{tabular}
    \label{runtime}
\end{center}
\end{table}

\section{Conclusion}

To fill the gap in existing navigation methods in FDR scenes, we propose a loop closure scheme that utilizes scene text cues inspired by human navigation. Our method fuses LiDAR and visual information to observe text entities in the environment and identify the authenticity of candidate loop closures through a graph-theoretic scheme. We collected multiple datasets in FDR scenarios and conducted comprehensive comparative experiments to demonstrate the competitiveness of our method. 
Our open-source code and dataset will be available for the community's benefit.

\bibliographystyle{IEEEtran}
\bibliography{main_RAL}

\end{document}